\pdfoutput=1
\documentclass[11pt]{article}
\usepackage{amsmath, amssymb}
\usepackage[final]{acl}
\usepackage{natbib} 
\usepackage{times}
\usepackage{latexsym}
\usepackage{multirow}
\usepackage{graphicx} 

\usepackage[T1]{fontenc}

\usepackage[utf8]{inputenc}
\usepackage{amsmath}
\usepackage{booktabs}
\usepackage{algorithm}
\usepackage{algpseudocode}

\usepackage{microtype}

\usepackage{inconsolata}

\usepackage{graphicx}

\title{Causal Invariance and Counterfactual Learning Driven Cooperative Game for Multi-Label Classification}

\author{
  \textbf{Yijia Fan$^{1,*}$, Jusheng Zhang$^{1,*}$, Kaitong Cai$^{1}$, Jing Yang$^{1}$, Keze Wang$^{1,\dagger}$}\\
  \vspace{2em}
  $^{1}$Sun Yat-sen University\\
  $^{\dagger}$Corresponding author: \texttt{kezewang@gmail.com}\\
  $^*$Equal contribution
}

\begin{document}
\maketitle
\begin{abstract}
Multi-label classification (MLC) faces persistent challenges from label imbalance, spurious correlations, and distribution shifts, especially in rare label prediction. We propose the Causal Cooperative Game (CCG) framework, which models MLC as a multi-player cooperative process. CCG integrates explicit causal discovery via Neural Structural Equation Models, a counterfactual curiosity reward to guide robust feature learning, and a causal invariance loss to ensure generalization across environments, along with targeted rare label enhancement. Extensive experiments on benchmark datasets demonstrate that CCG significantly improves rare label prediction and overall robustness compared to strong baselines. Ablation and qualitative analyses further validate the effectiveness and interpretability of each component. Our work highlights the promise of combining causal inference and cooperative game theory for more robust and interpretable multi-label learning.
\end{abstract}

\section{Introduction}

Multi-label classification (MLC) \cite{MLC,MLC2,MLC3,MLC4,MLC5} is a key task in machine learning \cite{mitchell1997machine}, widely applied in fields such as NLP. However, \cite{zongshu,zongshu2,Z1,f1}, real-world datasets often suffer from label imbalance, where rare labels have low representation in the training data \cite{zongshu3}. As a result, models tend to overlook these rare labels during training \cite{hanjian,hanjian2}, impacting prediction performance and generalization. As task complexity increases, effectively handling rare labels and improving model performance on imbalanced data remain significant challenges \cite{sun2009classification}.

Mainstream multi-label classification methods rely on the statistical correlations of labels, such as resampling or adjusting loss functions to enhance focus on rare labels \cite{pingheng,cui2019class}. However, these methods generally assume that labels are independently and identically distributed, failing to capture complex causal relationships, especially dependencies between rare labels and other labels \cite{lin2017focal}. Existing methods are primarily based on surface co-occurrence information, making it difficult to identify spurious correlations (e.g., the co-occurrence of high-frequency and rare labels \cite{tarekegn2021review}). This leads to insufficient generalization in rare label prediction. For instance, when the co-occurrence of label A and label B is merely a surface statistical relationship rather than a causal one, the model might incorrectly use such relationships for prediction, resulting in inaccurate outcomes \cite{henning2022survey}. Therefore, in environments with distribution shifts, reducing the impact of spurious correlations and enhancing model robustness becomes a significant challenge in multi-label classification \cite{huang2021balancing,read2019classifier,f2}. Furthermore, distribution shifts (e.g., inconsistencies between training and testing data distributions) further weaken the generalization ability of traditional models. Specifically, in rare label prediction, models often overly rely on features of frequent labels, neglecting the uniqueness of rare labels, which leads to performance degradation \cite{zhang2023deep,yang2022survey}. Hence, constructing new methods that can capture causal relationships among labels and improve the prediction accuracy of rare labels has become a key research direction.

To alleviate the rare label problem, many methods have proposed different strategies \cite{dealvis2024survey,zongshu,zongshu2,zongshu3}. Some methods balance label frequency through resampling techniques \cite{zhang2014review} or increase the training weight of rare labels by designing weighted loss functions \cite{ruder2017overview,jain2016large,Z3,f3}. Another set of methods \cite{zhang-etal-2018-multi} enhances the prediction ability of rare labels via multi-task learning or label embedding. Although these approaches have shown improvements, they still rely on surface statistical correlations and lack the modeling of potential causal relationships. In recent years, causal reasoning has gained attention in machine learning as a means to eliminate spurious correlations among labels and improve model robustness under distribution shifts. However, existing studies mainly focus on single-label causal modeling, and the exploration of causal relationships in multi-label tasks remains insufficient \cite{huang2016survey}.

The motivation of this study arises from the current shortcomings of multi-label classification methods in handling rare labels, particularly the insufficient utilization of causal relationships among labels, which causes models to be susceptible to spurious correlations \cite{11,22222,f4}. To address this challenge, we introduce the concept of causal reasoning. The goal of this research is to propose a novel causal cooperative game learning framework by modeling multi-label classification as a multi-player cooperative game process. In this framework, each player is responsible for a specific subset of labels and learns the real dependencies among labels through causal discovery methods. Specifically, the main innovations of this study include:
\begin{enumerate}
\setlength\itemindent{-0em} 
\item Designing a causal discovery module based on Neural Structural Equation Models (Neural SEM) to construct a dynamic causal graph among labels, thereby revealing the true causal dependency structure.
\item Proposing a counterfactual curiosity reward mechanism that generates counterfactual samples and compares predictions before and after interventions. This mechanism guides the model to focus on real causal features rather than surface statistical features.
\item Introducing a confounder adjustment strategy by incorporating a causal invariance loss, ensuring consistent predictions of causal labels across different environments.
\end{enumerate}
\section{Related Work}
\paragraph{Multi-Label Classification (MLC)}
Multi-label classification (MLC) \cite{zongshu,cui2019class,zongshu3,MLC,zm1} is a fundamental task in machine learning with widespread applications in natural language processing (NLP) \cite{zongshu,zongshu2,han2025debatetodetectreformulatingmisinformationdetection,LastingBench}, computer vision, and bioinformatics. Traditional MLC methods primarily rely on statistical correlations between labels, employing techniques such as over-sampling or weighted loss functions to enhance the learning of rare labels. However, these approaches \cite{pingheng,cui2019class,lin2017focal} often assume independent and identically distributed (i.i.d.) labels\cite{zhang2014review}, neglecting complex causal dependencies among them. This limitation becomes particularly problematic when dealing with label imbalance and distribution shifts, as existing methods often fail to capture the distinct characteristics of rare labels, leading to poor predictive performance. Thus, a key challenge in MLC research is effectively modeling complex causal dependencies between labels to improve the prediction accuracy of rare labels.
\paragraph{The Rare Label Problem}
One of the biggest challenges in MLC is the rare label problem, especially when datasets exhibit severe label imbalance\cite{pingheng,22222,11}. Traditional learning algorithms often struggle with rare label prediction due to their low frequency and insufficient training samples\cite{232323,li2025lion}. To address this issue, researchers have proposed various solutions, including resampling techniques and weighted loss functions that increase the training weight of rare labels\cite{8953804}. Additionally, multi-task learning and label embedding techniques have been explored to enhance rare label representation learning\cite{ruder2017overviewmultitasklearningdeep}. However, most of these methods rely on surface-level statistical relationships and fail to model the underlying causal dependencies among labels. Since causal relationships provide deeper insights into label interactions, ignoring them can lead to spurious correlations, ultimately reducing prediction accuracy.

\paragraph{Causal Machine Learning}
Causal inference has recently attracted attention for reducing spurious correlations and improving robustness in machine learning\cite{ruder2017overviewmultitasklearningdeep,renwu,yu2014large-scale,lan2025mcbe}. While effective in single-label tasks, its use in multi-label classification is still limited, with most existing work focusing only on simple label relationships. To address this, we propose a Neural SEM-based framework that builds dynamic causal graphs to better capture true label dependencies and improve rare label prediction, offering a novel approach by combining causal reasoning with cooperative game theory.

\section{Method}
Causal reasoning and cooperative game theory are applied to solve multi-label text classification challenges. We propose four innovations: Causal Structure Modeling: Using Neural SEM to construct label dependencies as a learnable causal graph\cite{feder2022causal,rozemberczki2022shapley,lan2025mcbee,Z2} $\mathcal{G} = (\mathcal{L}, \mathcal{E})$, capturing genuine dependencies via edges $e_{ij} \in \mathcal{E}$ while avoiding spurious correlations. Counterfactual Learning: Designing a curiosity reward $C_k(\mathbf{x})$ based on counterfactual reasoning, guiding the model to focus on causal features by comparing counterfactual samples\cite{louizos2017causal}. Invariance Principle: Introducing a causal invariance-based objective $\mathcal{L}_{\text{inv}}$ to ensure stable feature extraction across environments. Rare Label Enhancement: Using dynamic weights $w_{\text{rare}}(\ell)$ and a specialized loss function $\mathcal{L}_{\text{rare}}$ to improve rare label prediction. These innovations advance multi-label classification and causal learning in NLP. Subsequent chapters detail implementation, theoretical derivations, and experiments.
\subsection{Causality-Driven Multi-Label Cooperative Game Framework}
In this framework, the label prediction function is one of the core components, primarily responsible for capturing and modeling causal relationships between labels using the Neural SEM model. It is formally defined as follows:
Given a label set $\mathcal{L} = {\ell_1, \ell_2, \dots, \ell_L}$ and an input text feature representation $\mathbf{x} \in \mathbb{R}^d$ , the label prediction function $\widehat{y}_i$ predicts the probability of the $i$-th label $\ell_i$, defined as:
\scalebox{0.75}{$
\widehat{y}_i = \sigma \left( \sum_{\substack{j=1 \\ j \neq i}}^{L} w_{ij}^{(1)} \cdot h_{ij}^{(1)} \left( \mathbf{x}; \theta_{ij}^{(1)} \right) + b_i^{(1)} \right), \quad \forall i \in \{1, 2, \dots, L\}
$}
where $h_{ij}^{(1)}(\mathbf{x}; \theta_{ij}^{(1)})$ is a function learned using the Neural SEM model, capturing the causal relationship between the input features $\mathbf{x}$ and the labels; $w_{ij}^{(1)}$ represents the learned causal weight, indicating the influence of label $\ell_j$ on label $\ell_i$; $\theta_{ij}^{(1)}$ and $b_i^{(1)}$ are model parameters, including the neural network weights and bias; $\sigma(\cdot)$ is the sigmoid function, mapping the output to probabilities.
\subsection{Causal Graph Construction and Prior Constraints}
To construct the causal relationship graph among labels, we define a directed graph $\mathcal{G}$ based directly on weights and thresholds. This graph consists of a vertex set $\mathcal{L}$ and an edge set $\mathcal{E}$:
\scalebox{0.8}{$
\mathcal{G} = \left( \mathcal{L}, \underbrace{\left\{ (\ell_j \rightarrow \ell_i) \mid w_{ij}^{(1)} > \tau_{ij} \right\}}_{\mathcal{E}} \right), \quad \tau_{ij} = \Phi(\alpha_{ij}, \beta_{ij}) > 0
$}
where $w_{ij}^{(1)}$ represents the causal strength from label $\ell_j$ to label $\ell_i$, and $\tau_{ij}$ is the threshold determined by the function $\Phi$ based on parameters $\alpha_{ij}$ and $\beta_{ij}$. When the causal strength $w_{ij}^{(1)}$ exceeds the corresponding threshold $\tau_{ij}$, a directed edge $\ell_j \rightarrow \ell_i$ is established in the graph, indicating a significant causal influence.
This construction method determines causal relationships directly by comparing weights with thresholds, making the graph structure more interpretable and practically meaningful.
\subsection{Prior Constraints}
To address the rare label (low-frequency label) problem, we introduce a rare label indicator function $\mathcal{I}_{\text{rare}}(i,j)$ and a regulation operator $\Psi(\cdot)$ to enhance the causal edge weights for rare labels:
\scalebox{0.85}{$
\mathcal{I}_{\text{rare}}(i,j) = 
\mathbf{1} \left( \ell_i \in \mathcal{L}_{\text{rare}} \lor \ell_j \in \mathcal{L}_{\text{rare}} \right),
\quad 
\Psi(\eta) = \eta^{\mathcal{I}_{\text{rare}}(i,j)}
$}
where $\mathcal{L}_{\text{rare}} \subseteq \mathcal{L}$ denotes the set of low-frequency rare labels. If at least one of the labels in a given label pair is a rare label, the causal weight is amplified by a causal enhancement factor $\eta > 1$, ensuring that causal relationships involving rare labels are effectively captured.
\subsection{Causal Graph Learning Objective}
The goal of constructing the causal graph is to learn a weight matrix $w_{ij}^{(1)}$ that accurately captures the causal relationships between labels, making it as close as possible to the ideal causal weight $\tilde{w}_{ij}$. Based on this, we define the optimization objective for causal graph learning as follows:

\scalebox{0.82}{$
\begin{aligned}
\mathcal{L}_{\text{causal}} =\;& 
\sum_{i \neq j} 
\underbrace{\Psi(\eta)}_{\text{Rare Edge Enhancement}}
\cdot \left| w_{ij}^{(1)} - \tilde{w}_{ij} \right|_2^2 \\
&+ \lambda \cdot \sum_{i=j} 
\underbrace{| w_{ij}^{(1)} |_0}_{\text{Self-loop Suppression}} \\
\text{where}\quad
\tilde{w}_{ij} =\;& 
\underbrace{
\gamma \cdot f_{\text{co-occur}}(i,j) + (1-\gamma) \cdot f_{\text{semantic}}(i,j)
}_{\text{Ideal Weight Estimation}}
\end{aligned}
$}

The objective is to make the learned causal weight $w_{ij}^{(1)}$ closely approximate the ideal causal weight $\tilde{w}_{ij}$. The estimation of $\tilde{w}_{ij}$ is based on two weighted factors:$f_{\text{co-occur}}(i,j)$: The co-occurrence frequency of labels $i$ and $j$ in the dataset.$f_{\text{semantic}}(i,j)$: The semantic similarity between labels $i$ and $j$.The hyperparameter $\gamma \in [0,1]$ controls the relative importance of these two factors. The function $\Psi(\eta)$ applies to the entire term, making weight changes for rare label-related edges contribute more significantly to the loss, thereby enhancing the learning of causal relationships for rare labels.
Since causal relationships should reflect cross-label influences, we aim to avoid learning self-loops (i.e., causal edges of the form $\ell_i \to \ell_i$). To achieve this, we use the $\ell_0$ norm $|\cdot|_0$ to count the number of nonzero elements on the diagonal of the weight matrix. A regularization term with hyperparameter $\lambda$ is introduced to suppress self-loops, ensuring that the final learned causal graph does not contain excessive self-loops.

   \begin{algorithm}[t]
   \caption{Causality-Driven Multi-Label Classification Framework}
   \begin{algorithmic}[1]
   \State \textbf{Input:} Text feature $\mathbf{x}$, label set $\mathcal{L}$
   \State \textbf{Construct} causal graph $\mathcal{G} = (\mathcal{L}, \mathcal{E})$ via Neural SEM
   \State \textbf{Estimate} ideal causal weights $\tilde{w}_{ij}$ using co-occurrence and semantic similarity
   \State \textbf{Learn} causal weights $w_{ij}^{(1)}$ by minimizing $\mathcal{L}_{\text{causal}}$
   \State \textbf{Enhance} rare label edges with $\Psi(\eta)$
   \State \textbf{Suppress} self-loops via regularization
   \State \textbf{Partition} $\mathcal{L}$ into causal subgraphs $\{\mathcal{L}_k\}$ for each player $P_k$
   \State \textbf{Apply} causal mask $\mathbf{M}_k$ to restrict each $P_k$'s attention
   \For{each player $P_k$}
       \State \textbf{Predict} labels using masked features
       \State \textbf{Compute} counterfactual curiosity reward $C_k(\mathbf{x})$
       \State \textbf{Update} model with invariance loss $\mathcal{L}_{\text{inv}}$ and weighted cross-entropy
   \EndFor
   \State \textbf{Output:} Multi-label predictions $\{\widehat{y}_i\}$
   \end{algorithmic}
   \end{algorithm}
   
\subsection{Player Decomposition and Causal Constraints}
To mitigate the interference of spurious statistical correlations between labels, we propose a causal decoupling player mechanism based on the label causal graph $\mathcal{G} = (\mathcal{L}, \mathcal{E})$. This mechanism consists of two key steps:
\textbf{Causally-driven partitioning} of the label set $\mathcal{L}$ into $N$ mutually exclusive subsets ${\mathcal{L}k}{k=1}^{N}$, where each subset corresponds to the perceptual domain of an independent player $P_k$.
Using a causal mask matrix $\mathbf{M}_k$ to constrain the attention scope of player $P_k$. 
Causal Subgraph Partitioning:Based on the topological structure of $\mathcal{G}$, the label set $\mathcal{L}$ is partitioned as $\mathcal{L} = \bigcup_{k=1}^N \mathcal{L}_k$, ensuring that $\mathcal{L}k \cap \mathcal{L}{k'} = \varnothing$ for any $k \neq k'$. Each subset $\mathcal{L}_k$ corresponds to a Maximal Connected Causal Subgraph, forming a complete causal chain $\mathcal{C}_k = {\ell^{(k)}1 \xrightarrow{\epsilon_1} \ell^{(k)}2 \xrightarrow{\epsilon_2} \cdots \xrightarrow{\epsilon{m_k-1}} \ell^{(k)}{m_k}}$, where $\epsilon_i$ represents the causal effect strength. For example, if $\mathcal{L}k = {\ell{\text{root}}^{(k)}, \ell_{\text{mid}}^{(k)}, \ell_{\text{leaf}}^{(k)}}$, the corresponding causal pathway is $\ell_{\text{root}}^{(k)} \Rightarrow \ell_{\text{mid}}^{(k)} \Rightarrow \ell_{\text{leaf}}^{(k)}$, where $\Rightarrow$ denotes a direct causal effect.
Causal Perception Constraint:For each player $P_k$, we define a binary causal mask matrix $\mathbf{M}_k \in {0,1}^{L \times L}$, where each element satisfies:
\[
m_{ij}^{(k)} =
\begin{cases} 
1, & \text{if } (\ell_j \rightarrow \ell_i) \in \mathcal{E} \text{ and } \{\ell_j, \ell_i\} \subseteq \mathcal{L}_k, \\
0, & \text{otherwise}.
\end{cases}
\]
This mask applies through the Hadamard product $\odot$ on the feature interaction matrix, restricting $P_k$'s perception strictly to its assigned causal subgraph $\mathcal{G}_k = (\mathcal{L}_k, \mathcal{E}_k)$, where $\mathcal{E}_k = \mathcal{E} \cap (\mathcal{L}_k \times \mathcal{L}_k)$. For example, when $\mathcal{L}k = {\ell^{(k)}{\text{root}}, \ell^{(k)}{\text{med}}, \ell^{(k)}{\text{leaf}}}$, only the causal path $\ell^{(k)}{\text{root}} \to \ell^{(k)}{\text{med}} \to \ell^{(k)}_{\text{leaf}}$ is retained in $\mathbf{M}_k$, effectively eliminating spurious statistical correlations by filtering out interactions where $\ell_j \notin \mathcal{L}_k$ or $(\ell_j, \ell_i) \notin \mathcal{E}_k$.

\subsection{Counterfactual Curiosity Reward Mechanism}
This section introduces a counterfactual curiosity mechanism to enhance causal learning via:Method Design: Constructing a causal intervention-driven reward function $\mathcal{R}_{\text{cf}}$, leveraging counterfactual generation and causal invariance measurement.
Feature Learning: Encouraging the model $f_\theta$ to capture causal invariant features $\mathcal{C}$ while suppressing spurious correlations $\mathcal{S}$.
Performance Enhancement: Improving robustness $\rho$ and generalization $\mathcal{G}$ in adversarial environments $\mathcal{E}_{\text{adv}}$.

\subsubsection{Counterfactual Consistency Reward}
For each player $P_k$, we measure its prediction consistency for a sub-label $\ell_c$ on both the original sample $\mathbf{x}$ and the counterfactual sample $\mathbf{x}_{\text{cf}}$ using the Jensen-Shannon (JS) divergence, defined as:
\scalebox{0.8}{$
C_k^{\text{cf}}(\mathbf{x}) = -\text{JS} \left( \pi_k(\mathbf{x})_{\ell_c} \,\middle\|\, \pi_k(\mathbf{x}^{\text{cf}})_{\ell_c} \right)
$}
The JS divergence ranges in $[0, \log 2]$, ensuring symmetry and robustness to zero probabilities. The negative sign ensures that a smaller distribution difference results in a higher reward, promoting counterfactual stability. The overall reward function for player $P_k$ is formulated as:
\scalebox{0.85}{$
\begin{aligned}
C_k(\mathbf{x}) &= 
\underbrace{\frac{1}{|\mathcal{L}_k|} \sum_{\ell \in \mathcal{L}_k} 
\frac{\mathbf{1}\{y_\ell = y_\ell\}}{1 + \text{freq}(\ell)}
}_{\text{Rare Label Accuracy}} \\
&\quad + \beta \cdot 
\underbrace{D\left( \pi_k(\mathbf{x})_{\ell}, \overline{\pi}_{-k}(\mathbf{x})_{\ell} \right) 
+ \gamma \cdot C_k^{\text{cf}}(\mathbf{x})}_{\text{Prediction Diversity and Counterfactual Consistency}}.
\end{aligned}
$}
where the rare label accuracy term ensures balanced rewards across different label frequencies, giving higher weight to rare labels via $\frac{1}{1 + \text{freq}(\ell)}$. The prediction diversity term, defined as the KL divergence between the player’s prediction distribution $\pi_k(\mathbf{x})_\ell$ and the average distribution of other players $\overline{\pi}{-k}(\mathbf{x})\ell$, encourages exploration of diverse prediction patterns. The counterfactual consistency term $C_k^{\text{cf}}(\mathbf{x})$ ensures that player $P_k$ remains stable under feature interventions, forcing the model to focus on causal features.

Initialization Strategy: Set $\beta = \gamma = 1$ initially and adjust dynamically during training—increase $\beta$ in early stages to encourage exploration and prediction diversity, and increase $\gamma$ in later stages to strengthen causal feature learning.

\subsection{Causal Invariance Loss Function}
To enhance generalization under distribution shifts or interventions, we ensure the model learns invariant causal features across environments. We create augmented environments $\mathcal{E}_1, \mathcal{E}_2, \dots, \mathcal{E}M$ via synonym replacement, sentence restructuring, and grammar modifications, and intervention environments $\mathcal{E}{\text{int}}$ by perturbing non-causal features (e.g., background info) based on the causal graph $\mathcal{G}$. Given input $\mathbf{x}$, its representation in $\mathcal{E}_m$ is $\mathbf{x}^{(m)}$. To enhance causal feature stability, we impose a dual invariance constraint to better capture true causal relationships.
(1) Causal Feature Contrastive Loss
To enforce causal feature consistency across environments, we define the contrastive invariance loss:
\[
\mathcal{L}_{\text{inv}} = \sum_{1 \leq m < n \leq M} 
\left\| \mathbf{h}_k\left( \mathbf{x}^{(m)} \right) - \mathbf{h}_k\left( \mathbf{x}^{(n)} \right) \right\|_2^2
\]
where $M$ is the total number of environments, $\mathbf{x}^{(m)}$ is the input under $\mathcal{E}_m$, and $\mathbf{h}_k: \mathcal{X} \to \mathbb{R}^d$ is the causal feature encoder for player $P_k$. This loss ensures causal representations remain consistent across environments, preventing reliance on spurious correlations; (2) Cross-Environment Prediction Consistency Loss
To enforce alignment of sub-label predictions across environments, we define the loss function:
\[
\mathcal{L}_{\text{causal}} = \frac{1}{M} \sum_{m=1}^M 
\underbrace{\mathcal{H}\left( \mathbf{y}_{\mathcal{L}_k}, \pi_k(\mathbf{x}^{(m)})_{\mathcal{L}_k} \right)}_{\text{Cross-Entropy Loss}}
\]
where $\pi_k: \mathcal{X} \to \Delta^{|\mathcal{L}_k|}$ is the label prediction function for player $P_k$, $\mathbf{y}_{\mathcal{L}_k}$ denotes the true label distribution for sub-label set $\mathcal{L}_k$, and $\mathcal{H}(\cdot, \cdot)$ represents the cross-entropy function.
\subsection{Weighted Cross-Entropy Loss for Multi-Label Classification}
In multi-label classification, label imbalance causes varying learning difficulty between common and rare labels. To maintain recognition of common labels while enhancing rare label learning, we propose a \textbf{dual-supervision composite loss} with a dynamic weighting mechanism. It combines \textbf{weighted cross-entropy loss} and a \textbf{rare-label regularization} term. The basic form is:
\[
\mathcal{L}_{\text{base}} = -\frac{1}{N} \sum_{n=1}^N \sum_{\ell=1}^L 
\underbrace{\left[ \alpha(\ell) \cdot y_{n\ell} \log \sigma(\mathbf{h}_n^\top \mathbf{W}_\ell) \right]}_{\text{Weighted Cross-Entropy Loss}}
\]
where $\mathbf{h}_n \in \mathbb{R}^k$ is the hidden representation of sample $\mathbf{x}_n$, and $\mathbf{W}_\ell \in \mathbb{R}^k$ is the classification weight vector for label $\ell$, representing its feature representation. $\sigma(\cdot)$ is the Sigmoid activation function, mapping inputs to $[0,1]$, indicating the predicted probability of each label. $y_{n\ell} \in {0,1}$ is the ground truth for label $\ell$ on sample $n$.
The dynamic weighting factor $\alpha(\ell)$ adjusts the importance of each label in the overall loss function—lower for common labels and higher for rare labels, enhancing rare label learning.

\section{Experiments}

We conduct a series of experiments to comprehensively evaluate our proposed CCG framework, including: (1) comparative performance with baselines (\ref{sec:comparative_performance}), (2) qualitative causal analysis and visualization (\ref{sec:qualitative_analysis}), (3) analysis of the impact of player number (\ref{sec:player_analysis}), (4) ablation study (\ref{sec:ablation_study}), and (5) robustness to distribution shifts (\ref{sec:robustness_distribution_shift}). Unless otherwise specified, the number of players is set to $N=5$. Detailed hyperparameter settings for each experiment are provided in the supplementary material (Appendix~\ref{sec:appendix_parameter_settings}).

\subsection{Comparative Performance}
\label{sec:comparative_performance}

\paragraph{Experimental Setup}
To rigorously assess the efficacy of our proposed Causal Cooperative Game (CCG) framework, particularly its capability in addressing the critical challenge of rare label prediction in multi-label classification (MLC), we conduct comprehensive comparative experiments. The evaluation is performed on four widely recognized multi-label text classification benchmarks: \textbf{20 Newsgroups\cite{Newsweeder}}, \textbf{DBpedia\cite{DBpedia}}, \textbf{Ohsumed\cite{OHSUMED}}, and \textbf{Reuters news\cite{RCV1}}, which span various domains and exhibit different label distribution characteristics. We compare our method with representative baselines, including RoBERTa\cite{RoBERTa} (pre-trained language model) and several graph neural network-based methods: HGAT\cite{HGAT}, HyperGAT\cite{HyperGAT}, TextGCN\cite{TextGCN}, and DADGNN\cite{DADGNN}; And also TextING, another competitive model in this domain. For all experiments, we adhere to standard dataset splits and preprocessing protocols commonly used for these benchmarks to ensure a fair comparison. Performance is primarily evaluated using two key metrics: 1) \textbf{mAP (mean Average Precision\cite{MAP})}, a standard holistic measure for MLC tasks, and 2) \textbf{Rare-Label F1\cite{van1979information,charte2015addressing}}, which specifically focuses on the F1-score for subsets of labels with frequencies in the bottom $p\%$ (e.g., $p=20\%, 30\%, 40\%, 50\%$, as reported in Table~\ref{tab:rare_f1_comparison}), directly reflecting a model's proficiency in handling infrequent labels.

\begin{table*}[ht]
\centering
\caption{Comparison of Rare - F1 Metrics of Different Methods on Various Datasets}
\label{tab:rare_f1_comparison}
\scalebox{0.58}{ 
\begin{tabular}{lcccccccc}
\toprule
\multirow{2}{*}{Method} & \multicolumn{2}{c}{20 Newsgroups} & \multicolumn{2}{c}{DBpedia} & \multicolumn{2}{c}{Ohsumed} & \multicolumn{2}{c}{Reuters news} \\
\cmidrule(r){2 - 3} \cmidrule(r){4 - 5} \cmidrule(r){6 - 7} \cmidrule(r){8 - 9}
 & Rare - F1 @ 30\% & Rare - F1 @ 50\% & Rare - F1 @ 30\% & Rare - F1 @ 40\% & Rare - F1 @ 40\% & Rare - F1 @ 50\% & Rare - F1 @ 40\% & Rare - F1 @ 50\% \\
\midrule
RoBERTa & 75.3 & 65.4 & 55.4 & 41.4 & 47.6 & 41.4 & 47.2 & 43.4 \\
HGAT & 68.4 & 61.3 & 59.7 & 50.3 & 59.3 & 55.6 & 56.9 & 51.0 \\
HyperGAT & 69.3 & 62.4 & 60.2 & 51.4 & 60.4 & 56.1 & 58.1 & 53.6 \\
TextGCN & 67.2 & 61.8 & 57.4 & 48.3 & 57.3 & 51.3 & 54.7 & 51.4 \\
DADGNN & 72.2 & 62.1 & 61.3 & 51.9 & 61.2 & 57.4 & 59.6 & 54.1 \\
TextING & 74.6 & 64.8 & 62.4 & 52.4 & 62.5 & 58.2 & 61.3 & 55.8 \\
ours & 76.1 & 66.2 & 62.9 & 52.9 & 63.4 & 59.3 & 62.9 & 56.7 \\
\bottomrule
\end{tabular}
}
\end{table*}
\paragraph{Experimental Results}
The comparative performance of our CCG framework against baseline methods is detailed in Table~\ref{tab:rare_f1_comparison}. Across all four benchmark datasets and various Rare-Label F1 thresholds, our proposed method (``ours'') consistently demonstrates superior or highly competitive performance. For instance, on the DBpedia dataset, ``ours'' achieves a Rare-F1@30\% of 62.9 and Rare-F1@40\% of 52.9, outperforming the strongest baselines such as TextING (62.4 and 52.4, respectively). Similar advantages are observed on 20 Newsgroups (e.g., ``ours'' with 76.1 versus RoBERTa with 75.3 at Rare-F1@30\%), Ohsumed (e.g., ``ours'' with 63.4 versus TextING with 62.5 at Rare-F1@40\%), and Reuters news (e.g., ``ours'' with 62.9 versus TextING with 61.3 at Rare-F1@40\%). This consistent and significant improvement in Rare-Label F1 scores underscores the efficacy of our framework in mitigating the label imbalance problem and enhancing the recognition of underrepresented categories. By moving beyond potentially spurious statistical correlations that often mislead models, especially in the context of rare labels, our CCG approach demonstrates a notable advancement in building more robust and accurate multi-label classification systems.
\subsection{Deeper Causal Analysis and Visualization}
\label{sec:qualitative_analysis}
\paragraph{Experimental Setup}
Quantitative metrics (Section \ref{sec:comparative_performance}) summarize performance but don't fully reveal inter-label dependencies learned by our Causal Cooperative Game (CCG) framework. To capture genuine, potentially causal relationships in the learned causal graph $\mathcal{G} = (\mathcal{L}, \mathcal{E})$, we conduct qualitative analysis on the \textbf{Ohsumed} dataset, leveraging its rich medical label semantics for intuitive relational validity assessment. We visualize subgraphs of $\mathcal{G}$ by thresholding causal weights $w_{ij}^{(1)}$, focusing on subgraphs with common and rare labels or varying clinical specificity. These are evaluated for coherence with medical knowledge or logical consistency, assessing if the model learns meaningful mechanisms rather than superficial correlations.
\begin{figure}
    \centering
    \includegraphics[width=\linewidth]{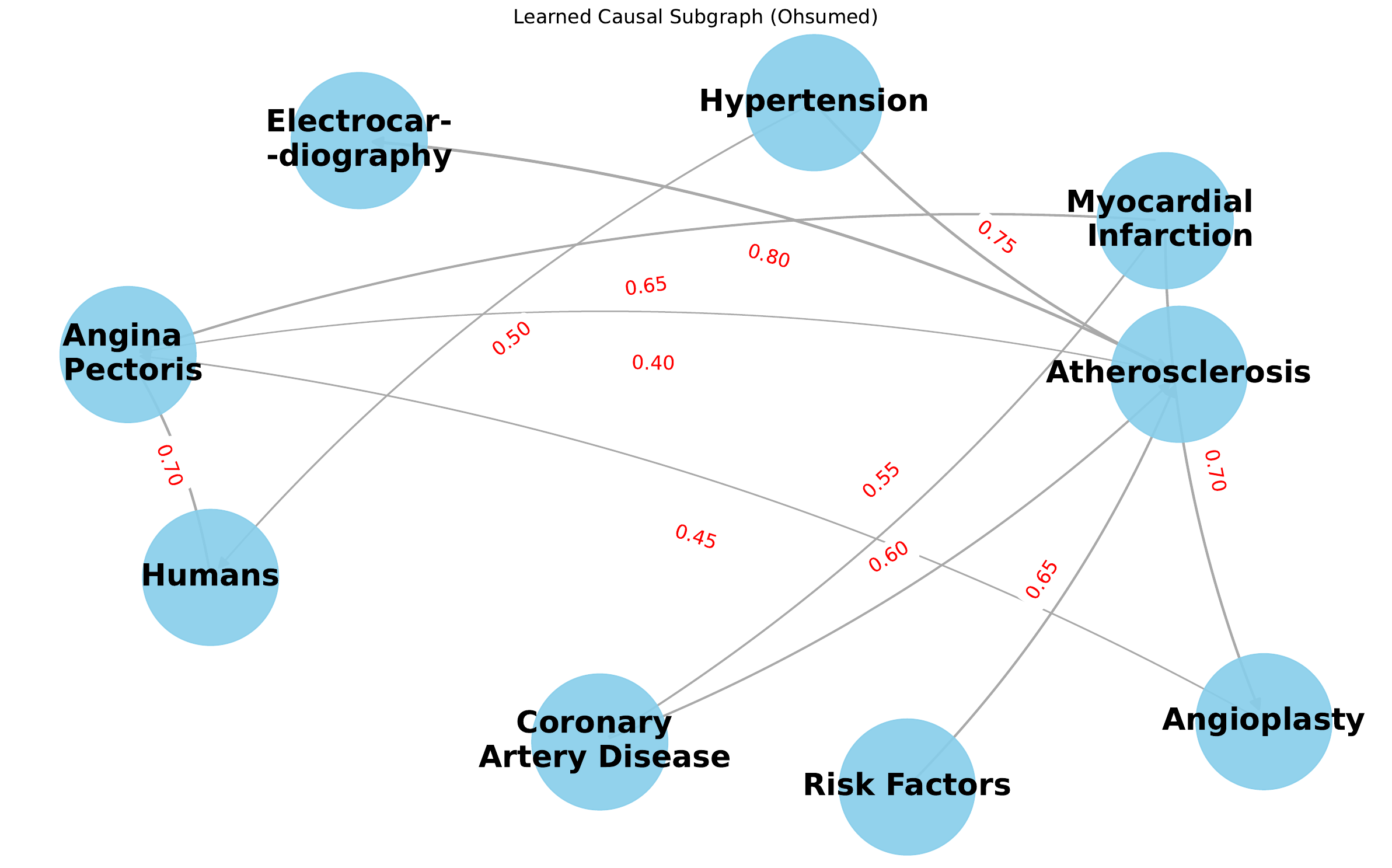}
    \caption{A subgraph showing the learned (hypothesized) causal relationships between concepts related to cardiovascular disease. The nodes represent specific medical labels, and the edges and their accompanying red weights indicate the mutual influence and learned strength between these concepts.}
    \label{fig:shiyan2}
\end{figure}
\paragraph{Experimental Results}
As shown in Figure~\ref{fig:shiyan2}, the model captures clinically meaningful multi-step dependencies in the cardiovascular domain. For example, “Humans” → “Risk Factors” ($w=0.80$) → “Hypertension” ($w=0.70$) and “Atherosclerosis” ($w=0.60$); “Atherosclerosis” → “Coronary Artery Disease” ($w=0.85$) → “Angina Pectoris” ($w=0.70$) and “Myocardial Infarction” ($w=0.80$); and “Myocardial Infarction” → “Electrocardiography” ($w=0.90$) and “Angioplasty” ($w=0.80$). These results demonstrate that our CCG framework learns interpretable, clinically relevant multi-step relationships rather than mere surface associations.
\subsection{Analysis of Player Number Impact in Cooperative Game Framework}
\label{sec:player_analysis}
\paragraph{Experimental Setup}
A key feature of our Causal Cooperative Game (CCG) framework is partitioning the label set $\mathcal{L}$ into $N$ disjoint subsets, each handled by an independent player $P_k$. The choice of $N$ affects how well local causal dependencies are captured and spurious correlations are reduced: too small $N$ may weaken causal decoupling, while too large $N$ may fragment meaningful causal chains. To assess the sensitivity of CCG to this hyperparameter, we vary $N$ (1, 2, 3, 4, 5, 6, 8, 10) on the \textbf{20 Newsgroups} dataset, using our causal subgraph partitioning for $N>1$ and assigning all labels to one player for $N=1$. All other settings are fixed, and performance is measured by mAP and F1-score.
\begin{figure}
    \centering
    \includegraphics[width=\linewidth]{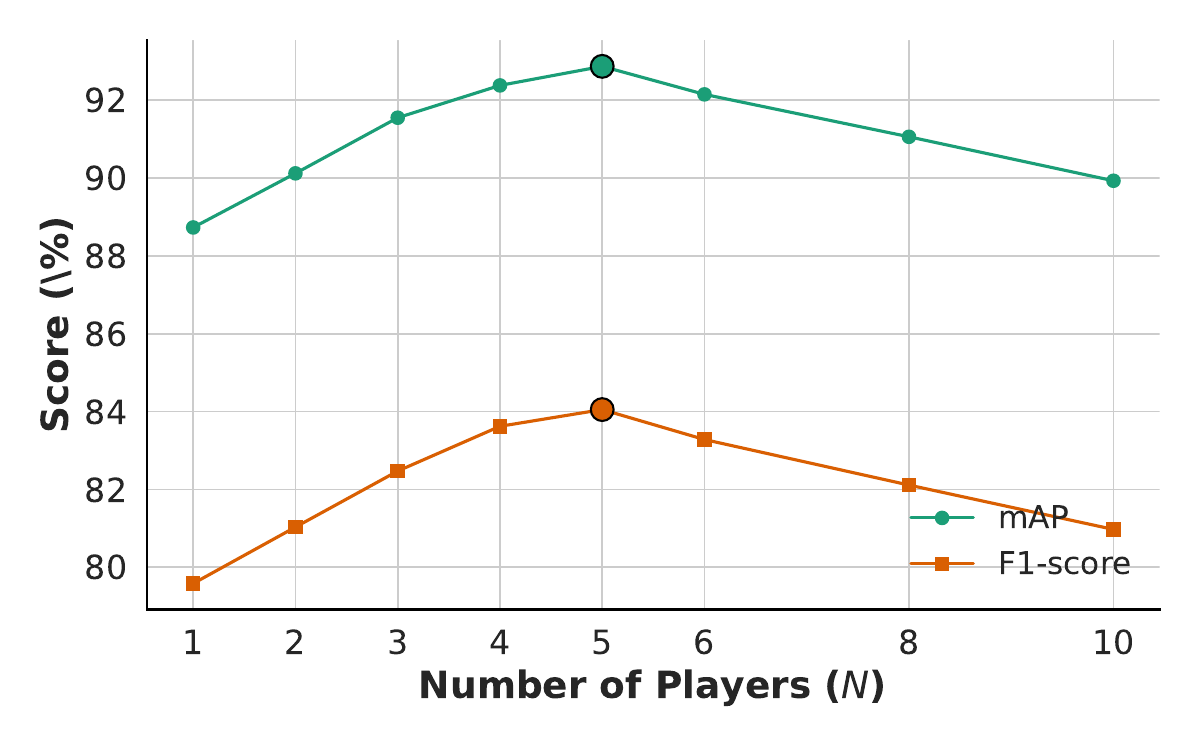}
    \caption{The curve of mAP score and F1-score of CCG on 20 Newsgroups with the change of Number of Players}
    \label{fig:fake_player_impact}
\end{figure}
\paragraph{Experimental Results}
Figure~\ref{fig:fake_player_impact} shows that increasing the number of players ($N$) on the 20 Newsgroups dataset initially improves both mAP and F1-score, peaking at $N=5$ (mAP 92.87\%, F1-score 84.05\%). With $N=1$, the model is less effective, lacking the benefits of causal decoupling. Performance gains up to $N=5$ suggest that moderate partitioning enables each player to better capture local dependencies and reduce spurious correlations. However, further increasing $N$ leads to a decline, likely due to over-fragmentation and loss of broader causal context. These results highlight the importance of choosing an appropriate $N$ to balance granularity and context in our CCG framework.
\begin{table*}[t]
\centering
\caption{Ablation study results on the DBpedia dataset, showing the impact of removing key components from our Full Causal Cooperative Game (CCG) model. Performance is reported in terms of mAP (\%) and Rare-Label F1 (\%). Best performance is highlighted in \textbf{bold}.}
\label{tab:fake_ablation_study}
\scalebox{0.85}{  
\begin{tabular}{lcc}
\toprule
Model Variant & DBpedia - mAP & DBpedia - Rare-Label F1 \\
\midrule
Full Model (CCG)                       & \textbf{89.15} & \textbf{78.23}   \\
\midrule
w/o Causal Graph Modeling (CGM)        & 87.58 & 76.17   \\
w/o Counterfactual Curiosity Reward (CCR)& 86.72 & 75.06   \\
w/o Causal Invariance Loss (CIL)       & 87.31 & 75.84   \\
w/o Multi-Player Decomposition (MPD)   & 86.05 & 74.22   \\
w/o Rare Label Enhancement (RLE)       & 88.03 & 72.95   \\
\bottomrule
\end{tabular}
}
\end{table*}
\subsection{Ablation Study}
\label{sec:ablation_study}
\paragraph{Experimental Setup}
To evaluate the contributions of our Causal Cooperative Game (CCG) framework's components—Causal Graph Modeling (CGM), Counterfactual Curiosity Reward (CCR), Causal Invariance Loss (CIL), Multi-Player Decomposition (MPD), and Rare Label Enhancement (RLE)—we perform an ablation study on the \textbf{DBpedia} dataset, a standard multi-label classification benchmark. From the full CCG model, we remove one component at a time, keeping model architecture and hyperparameters fixed. We assess each ablation’s impact using \textbf{mAP} and \textbf{Rare-Label F1} metrics to quantify contributions to rare label prediction and robust inter-label dependency learning.
\paragraph{Experimental Results}
Table~\ref{tab:fake_ablation_study} shows that the full CCG model achieves the best performance on DBpedia (89.15\% mAP, 78.23\% Rare-Label F1). Removing key components leads to clear drops: without Causal Graph Modeling (CGM), mAP and Rare-Label F1 fall to 87.58\% and 76.17\%; without Counterfactual Curiosity Reward (CCR), to 86.72\% and 75.06\%; and without Multi-Player Decomposition (MPD), to 86.05\% and 74.22\%. This highlights the importance of explicit causal structure, counterfactual guidance, and cooperative decomposition. Excluding Causal Invariance Loss (CIL) also reduces generalization (87.31\% mAP, 75.84\% Rare-Label F1). Notably, removing Rare Label Enhancement (RLE) most severely impacts rare label F1 (down to 72.95\%), confirming its effectiveness for label imbalance. Overall, each component synergistically improves rare label prediction and robust modeling of true inter-label dependencies.
\begin{figure}
    \centering
    \includegraphics[width=\linewidth]{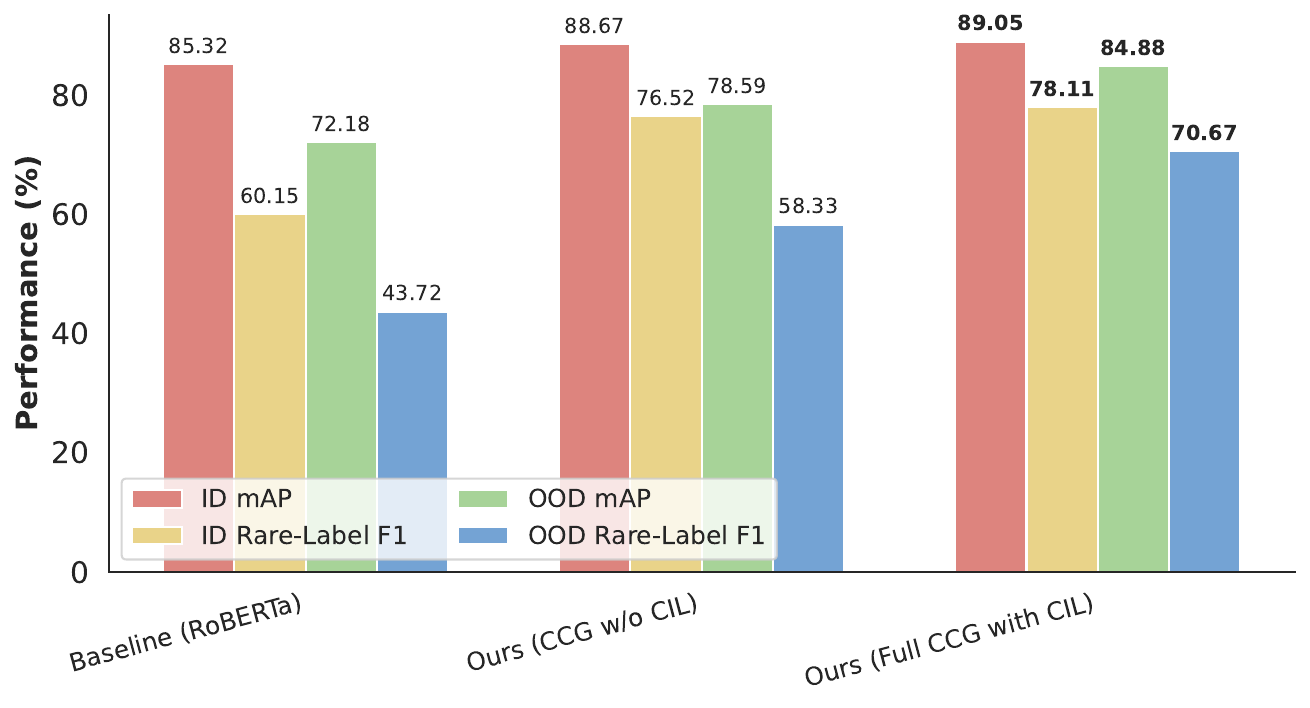}
    \caption{Performance comparison under simulated temporal distribution shift on the Reuters Corpus Volume 1 (RCV1) dataset. ID denotes In-Distribution test set (earlier period), and OOD denotes Out-of-Distribution test set (later period). $\Delta$ indicates the absolute performance drop from ID to OOD. Best OOD performance and smallest degradation are highlighted in \textbf{bold}.}
    \label{fig:fake_distribution_shift_results_rcv1}
\end{figure}
\subsection{Robustness to Distribution Shifts}
\label{sec:robustness_distribution_shift}
\paragraph{Experimental Setup}
Robustness to distribution shifts is crucial for real-world multi-label classification. To test this, we use the \textbf{Reuters Corpus Volume 1 (RCV1)} dataset and simulate a temporal shift by training on earlier articles and evaluating on both in-distribution (ID) and out-of-distribution (OOD, later period) test sets. This setup reflects realistic changes in topics and language over time. We compare a strong non-causal baseline (e.g., RoBERTa), our CCG without Causal Invariance Loss (CIL), and the full CCG model. All models are trained on the same data, and we report mAP and Rare-Label F1 to analyze generalization under distribution shift.


\paragraph{Experimental Results}
Figure~\ref{fig:fake_distribution_shift_results_rcv1} shows that all models experience performance drops on the OOD test set of RCV1. The baseline (RoBERTa) suffers large declines (mAP: -13.14\%, Rare-Label F1: -16.43\%), while CCG without CIL, though better on ID, still drops sharply on OOD, especially for rare labels. In contrast, our full CCG with CIL achieves the best ID results (89.05\% mAP, 78.11\% Rare-Label F1) and shows the smallest OOD degradation (mAP: -4.17\%, Rare-Label F1: -7.44\%). These results demonstrate that the CIL component enables our CCG framework to generalize better and maintain high predictive accuracy, even under significant distribution shifts inherent in real-world data streams like news articles. Overall, our model shows strong effectiveness in mitigating the challenges of robustness to distribution shifts.

\section{Conclusion}
\label{sec:conclusion}

This paper tackles key challenges in multi-label classification (MLC), particularly rare label prediction and spurious correlation mitigation, by introducing the Causal Cooperative Game (CCG) framework. CCG reformulates MLC as a multi-player cooperative process, combining explicit causal discovery with Neural SEMs, a counterfactual curiosity reward for robust feature learning, a causal invariance principle for stable predictions, and targeted rare label enhancement. Extensive experiments on benchmark datasets show that CCG notably improves performance—especially for rare labels—and enhances robustness to distribution shifts. Ablation studies confirm the importance of each component, while qualitative analysis demonstrates the interpretability of learned causal structures. This work points to a promising direction for building more robust, generalizable, and interpretable MLC systems through causal inference and cooperative game theory.

\section{Limitations and Future Works}
\label{sec:limitations_future_work}

One key area for future exploration and a current limitation of our Causal Cooperative Game (CCG) framework pertains to its player decomposition strategy. While the present causally-driven partitioning of labels is based on an initially learned causal graph structure and remains static throughout the training process, we identify this as an aspect with potential for enhancement. Future work will therefore investigate the development of more dynamic or adaptive player coalition formation mechanisms, which could potentially respond to evolving learned dependencies. Pursuing these research directions promises to further strengthen the capabilities of the CCG approach.

\section*{Acknowledgments}
This work was supported in part by the National Natural Science Foundation of China (NSFC) under Grant 62276283, in part by the China Meteorological Administration's Science and Technology Project under Grant CMAJBGS202517, in part by Guangdong Basic and Applied Basic Research Foundation under Grant 2023A1515012985, in part by Guangdong-Hong Kong-Macao Greater Bay Area Meteorological Technology Collaborative Research Project under Grant GHMA2024Z04, in part by Fundamental Research Funds for the Central Universities, Sun Yat-sen University under Grant 23hytd006, and in part by Guangdong Provincial High-Level Young Talent Program under Grant RL2024-151-2-11.

\appendix
\bibliography{custom}

@INPROCEEDINGS{MLC,
  author={Venkatesan, Rajasekar and Er, Meng Joo},
  title={Multi-label classification method based on extreme learning machines}, 
  year={2014},
  volume={},
  number={},
  pages={619-624},
  doi={10.1109/ICARCV.2014.7064375}}

@ARTICLE{MLC2,
  author={Zhang, Min-Ling and Zhou, Zhi-Hua},
  journal={IEEE Transactions on Knowledge and Data Engineering}, 
  title={A Review on Multi-Label Learning Algorithms}, 
  year={2014},
  volume={26},
  number={8},
  pages={1819-1837},
  doi={10.1109/TKDE.2013.39}}

@article{MLC3,
   title={Classifier Chains: A Review and Perspectives},
   volume={70},
   ISSN={1076-9757},
   url={http://dx.doi.org/10.1613/jair.1.12376},
   DOI={10.1613/jair.1.12376},
   journal={Journal of Artificial Intelligence Research},
   publisher={AI Access Foundation},
   author={Read, Jesse and Pfahringer, Bernhard and Holmes, Geoffrey and Frank, Eibe},
   year={2021},
   month=feb, pages={683–718} }

@ARTICLE{MLC4,
  author={Ghani, Muhammad Usman and Rafi, Muhammad and Tahir, Muhammad Atif},
  journal={IEEE Access}, 
  title={Discriminative Adaptive Sets for Multi-Label Classification}, 
  year={2020},
  volume={8},
  number={},
  pages={227579-227595},
  doi={10.1109/ACCESS.2020.3041763}}

@article{MLC5,
   title={The Emerging Trends of Multi-Label Learning},
   volume={44},
   ISSN={1939-3539},
   url={http://dx.doi.org/10.1109/TPAMI.2021.3119334},
   DOI={10.1109/tpami.2021.3119334},
   number={11},
   journal={IEEE Transactions on Pattern Analysis and Machine Intelligence},
   publisher={Institute of Electrical and Electronics Engineers (IEEE)},
   author={Liu, Weiwei and Wang, Haobo and Shen, Xiaobo and Tsang, Ivor W.},
   year={2022},
   month=nov, pages={7955–7974} }

@book{mitchell1997machine,
  title={Machine Learning},
  author={Mitchell, Tom M.},
  year={1997},
  publisher={McGraw-Hill}
}

@book{zongshu,
  title     = {Speech and Language Processing},
  author    = {Jurafsky, Daniel and Martin, James H.},
  year      = {2009},
  edition   = {2nd},
  publisher = {Prentice Hall}
}

@misc{zongshu2,
      title={Recent Trends in Deep Learning Based Natural Language Processing}, 
      author={Tom Young and Devamanyu Hazarika and Soujanya Poria and Erik Cambria},
      year={2018},
      eprint={1708.02709},
      archivePrefix={arXiv},
      primaryClass={cs.CL},
      url={https://arxiv.org/abs/1708.02709}, 
}

@ARTICLE{zongshu3,
  author={He, Haibo and Garcia, Edwardo A.},
  journal={IEEE Transactions on Knowledge and Data Engineering}, 
  title={Learning from Imbalanced Data}, 
  year={2009},
  volume={21},
  number={9},
  pages={1263-1284},
  doi={10.1109/TKDE.2008.239}}

@INPROCEEDINGS{hanjian,
  author={Spelmen, Vimalraj S and Porkodi, R},
  booktitle={2018 International Conference on Current Trends towards Converging Technologies (ICCTCT)}, 
  title={A Review on Handling Imbalanced Data}, 
  year={2018},
  volume={},
  number={},
  pages={1-11},
  doi={10.1109/ICCTCT.2018.8551020}}

@INBOOK{hanjian2,
  author={He, Haibo and Ma, Yunqian},
  booktitle={Imbalanced Learning: Foundations, Algorithms, and Applications}, 
  title={Foundations of Imbalanced Learning}, 
  year={2013},
  volume={},
  number={},
  pages={13-41},
  doi={10.1002/9781118646106.ch2}}

@article{sun2009classification,
  title   = {Classification of imbalanced data: A review},
  author  = {Sun, Yanmin and Wong, Andrew K.~C. and Kamel, Mohamed~S.},
  journal = {International Journal of Pattern Recognition and Artificial Intelligence},
  volume  = {23},
  number  = {4},
  pages   = {687--719},
  year    = {2009},
  publisher = {World Scientific},
  doi     = {10.1142/S0218001409007326}
}

@inproceedings{zm1,
author = {Wen, Zimo and Hu, Hanwen and Fang, Nan and Qian, Shiyou and Cao, Jian},
title = {DANet: A RAG-inspired Dual Attention Model for Few-shot Time Series Prediction},
year = {2025},
isbn = {9798400720406},
publisher = {Association for Computing Machinery},
address = {New York, NY, USA},
url = {https://doi.org/10.1145/3746252.3761012},
doi = {10.1145/3746252.3761012},
booktitle = {Proceedings of the 34th ACM International Conference on Information and Knowledge Management},
pages = {3302–3311},
numpages = {10},
location = {Seoul, Republic of Korea},
series = {CIKM '25}
}

@article{pingheng,
  title   = {Addressing Imbalance in Multi‐label Classification: Measures and Random Resampling Techniques},
  author  = {Charte, Francisco and Charte, David and Garc{\'\i}a, Salvador and Herrera, Fernando},
  journal = {Neurocomputing},
  volume  = {163},
  pages   = {3--16},
  year    = {2015},
  publisher = {Elsevier}
}

@inproceedings{zhang-etal-2018-multi,
    title = "Multi-Task Label Embedding for Text Classification",
    author = "Zhang, Honglun  and
      Xiao, Liqiang  and
      Chen, Wenqing  and
      Wang, Yongkun  and
      Jin, Yaohui",
    editor = "Riloff, Ellen  and
      Chiang, David  and
      Hockenmaier, Julia  and
      Tsujii, Jun{'}ichi",
    booktitle = "Proceedings of the 2018 Conference on Empirical Methods in Natural Language Processing",
    month = oct # "-" # nov,
    year = "2018",
    address = "Brussels, Belgium",
    publisher = "Association for Computational Linguistics",
    url = "https://aclanthology.org/D18-1484/",
    doi = "10.18653/v1/D18-1484",
    pages = "4545--4553",
}

@inproceedings{cui2019class,
  title     = {Class‐Balanced Loss Based on Effective Number of Samples},
  author    = {Cui, Yin and Jia, Menglin and Lin, Tsung‐Yi and Song, Yang and Belongie, Serge},
  booktitle = {Proceedings of the IEEE/CVF Conference on Computer Vision and Pattern Recognition (CVPR)},
  pages     = {9268--9277},
  year      = {2019}
}

@inproceedings{lin2017focal,
  title     = {Focal Loss for Dense Object Detection},
  author    = {Lin, Tsung‐Yi and Goyal, Priya and Girshick, Ross and He, Kaiming and Dollár, Piotr},
  booktitle = {Proceedings of the IEEE International Conference on Computer Vision (ICCV)},
  pages     = {2980--2988},
  year      = {2017}
}

@article{tarekegn2021review,
  title   = {A review of methods for imbalanced multi-label classification},
  author  = {Tarekegn, Adane Nega and Liu, Xixi and Li, Yong},
  journal = {OpenReview},
  year    = {2021},
  url     = {https://openreview.net/forum?id=ZB6muyOU6l}
}

@article{henning2022survey,
  title   = {A Survey of Methods for Addressing Class Imbalance in Deep-Learning Based Natural Language Processing},
  author  = {Henning, Sophie and Beluch, William and Fraser, Alexander and Friedrich, Annemarie},
  year    = {2022},
  archivePrefix = {arXiv},
  eprint  = {2210.04675},
  howpublished = {\\url{https://arxiv.org/abs/2210.04675}}
}

@inproceedings{huang2021balancing,
  title     = {Balancing Methods for Multi-label Text Classification with Long-Tailed Class Distribution},
  author    = {Huang, Yi and Giledereli, Buse and Köksal, Abdullatif and Özgür, Arzucan and Ozkirimli, Elif},
  booktitle = {arXiv preprint},
  year      = {2021},
  archivePrefix = {arXiv},
  eprint    = {2109.04712},
  howpublished = {\\url{https://arxiv.org/abs/2109.04712}}
}

@article{read2019classifier,
  title   = {Classifier Chains: A Review and Perspectives},
  author  = {Read, Jesse and Pfahringer, Bernhard and Holmes, Geoff and Frank, Eibe},
  journal = {arXiv preprint},
  year    = {2019},
  archivePrefix = {arXiv},
  eprint  = {1912.13405},
  howpublished = {\\url{https://arxiv.org/abs/1912.13405}}
}

@article{zhang2023deep,
  title   = {Deep Long-Tailed Learning: A Survey},
  author  = {Zhang, Yifan and Kang, Bingyi and Hooi, Bryan and Yan, Shuicheng and Feng, Jiashi},
  journal = {IEEE Transactions on Pattern Analysis and Machine Intelligence},
  year    = {2023},
  doi     = {10.1109/TPAMI.2023.3268118}
}

@article{yang2022survey,
  title   = {A Survey on Long-Tailed Visual Recognition},
  author  = {Yang, Lu and Jiang, He and Song, Qing and Guo, Jun},
  journal = {International Journal of Computer Vision},
  volume  = {130},
  number  = {9},
  pages   = {2150--2176},
  year    = {2022},
  doi     = {10.1007/s11263-022-01622-8}
}

@article{dealvis2024survey,
  title   = {A Survey of Deep Long-Tail Classification Advancements},
  author  = {de Alvis, Charika and Seneviratne, Suranga},
  journal = {arXiv preprint},
  year    = {2024},
  archivePrefix = {arXiv},
  eprint  = {2404.15593},
  howpublished = {\url{https://arxiv.org/abs/2404.15593}}
}

@article{zhang2014review,
  title   = {A Review on Multi-Label Learning},
  author  = {Zhang, Min-Ling and Zhou, Zhi-Hua},
  journal = {IEEE Transactions on Knowledge and Data Engineering},
  volume  = {26},
  number  = {8},
  pages   = {1819--1837},
  year    = {2014},
  doi     = {10.1109/TKDE.2013.39}
}

@article{ruder2017overview,
  title   = {An Overview of Multi‐Task Learning in Deep Neural Networks},
  author  = {Ruder, Sebastian},
  journal = {arXiv preprint},
  year    = {2017},
  archivePrefix = {arXiv},
  eprint  = {1706.05098},
  howpublished = {\\url{https://arxiv.org/abs/1706.05098}}
}

@inproceedings{jain2016large,
  title     = {Large‐Scale Multi‐Label Learning with Missing Labels},
  author    = {Jain, Prateek and Kulis, Brian and Dhillon, Inderjit S.},
  booktitle = {Proceedings of the 33rd International Conference on Machine Learning (ICML)},
  pages     = {593--602},
  year      = {2016}
}

@article{huang2016survey,
  title   = {A Survey of Causal Discovery and Causal Inference},
  author  = {Huang, Yuyang and Glymour, Clark},
  journal = {arXiv preprint},
  year    = {2016},
  archivePrefix = {arXiv},
  eprint  = {1611.06289},
  howpublished = {\url{https://arxiv.org/abs/1611.06289}}
}

@inproceedings{11,
  title     = {On the Consistency of Multi-Label Learning},
  author    = {Dembczy\'nski, Krzysztof and Waegeman, Willem and Cheng, Wei and Hüllermeier, Eyke},
  booktitle = {Proceedings of the 27th International Conference on Machine Learning (ICML)},
  pages     = {226–233},
  year      = {2010}
}

@ARTICLE{22222,
  author={Zhang, Min-Ling and Zhou, Zhi-Hua},
  journal={IEEE Transactions on Knowledge and Data Engineering}, 
  title={A Review on Multi-Label Learning Algorithms}, 
  year={2014},
  volume={26},
  number={8},
  pages={1819-1837},
  doi={10.1109/TKDE.2013.39}}

@article{232323,
author = {Buda, Mateusz and Maki, Atsuto and Mazurowski, Maciej A.},
title = {A systematic study of the class imbalance problem in convolutional neural networks},
year = {2018},
issue_date = {Oct 2018},
publisher = {Elsevier Science Ltd.},
address = {GBR},
volume = {106},
number = {C},
issn = {0893-6080},
url = {https://doi.org/10.1016/j.neunet.2018.07.011},
doi = {10.1016/j.neunet.2018.07.011},
journal = {Neural Netw.},
month = oct,
pages = {249–259},
numpages = {11}
}

@INPROCEEDINGS{8953804,
  author={Cui, Yin and Jia, Menglin and Lin, Tsung-Yi and Song, Yang and Belongie, Serge},
  booktitle={2019 IEEE/CVF Conference on Computer Vision and Pattern Recognition (CVPR)}, 
  title={Class-Balanced Loss Based on Effective Number of Samples}, 
  year={2019},
  volume={},
  number={},
  pages={9260-9269},
  doi={10.1109/CVPR.2019.00949}}

@misc{ruder2017overviewmultitasklearningdeep,
      title={An Overview of Multi-Task Learning in Deep Neural Networks}, 
      author={Sebastian Ruder},
      year={2017},
      eprint={1706.05098},
      archivePrefix={arXiv},
      primaryClass={cs.LG},
      url={https://arxiv.org/abs/1706.05098}, 
}

@misc{renwu,
      title={Multi-Task Learning with Deep Neural Networks: A Survey}, 
      author={Michael Crawshaw},
      year={2020},
      eprint={2009.09796},
      archivePrefix={arXiv},
      primaryClass={cs.LG},
      url={https://arxiv.org/abs/2009.09796}, 
}

@inproceedings{yu2014large-scale,
  title     = {Large-scale Multi-label Learning with Missing Labels},
  author    = {Yu, Hsiang-Fu and Jain, Prateek and Kar, Purushottam and Dhillon, Inderjit S.},
  booktitle = {Proceedings of the 31st International Conference on Machine Learning (ICML)},
  volume    = {32},
  pages     = {593--601},
  year      = {2014},
  publisher = {PMLR},
  url       = {https://proceedings.mlr.press/v32/yu14.html}
}

@article{feder2022causal,
  title     = {Causal Inference in Natural Language Processing: Estimation, Prediction, Interpretation and Beyond},
  author    = {Feder, Amir and Keith, Katherine A. and Manzoor, Emaad and Pryzant, Reid and Sridhar, Dhanya and Wood-Doughty, Zach and Eisenstein, Jacob and Grimmer, Justin and Reichart, Roi and Roberts, Margaret E. and Stewart, Brandon M. and Veitch, Victor and Yang, Diyi},
  journal   = {Transactions of the Association for Computational Linguistics},
  volume    = {10},
  pages     = {1138--1158},
  year      = {2022},
  doi       = {10.1162/tacl\_a\_00511}
}

@article{rozemberczki2022shapley,
  title   = {The Shapley Value in Machine Learning},
  author  = {Rozemberczki, Benedek and Watson, Lauren and Bayer, Péter and Yang, Hao-Tsung and Kiss, Olivér and Nilsson, Sebastian and Sarkar, Rik},
  journal = {arXiv preprint arXiv:2202.05594},
  year    = {2022},
  url     = {https://arxiv.org/abs/2202.05594}
}

@inproceedings{louizos2017causal,
  title     = {Causal Effect Inference with Deep Latent-Variable Models},
  author    = {Louizos, C. and Shalit, U. and Mooij, J. and Sontag, D. and Zemel, R. and Welling, M.},
  booktitle = {Proceedings of the 31st Conference on Neural Information Processing Systems (NeurIPS)},
  pages     = {6446--6456},
  year      = {2017}
}

@misc{RoBERTa,
      title={RoBERTa: A Robustly Optimized BERT Pretraining Approach}, 
      author={Yinhan Liu and Myle Ott and Naman Goyal and Jingfei Du and Mandar Joshi and Danqi Chen and Omer Levy and Mike Lewis and Luke Zettlemoyer and Veselin Stoyanov},
      year={2019},
      eprint={1907.11692},
      archivePrefix={arXiv},
      primaryClass={cs.CL},
      url={https://arxiv.org/abs/1907.11692}, 
}

@inproceedings{Newsweeder,
  title     = {Newsweeder: Learning to Filter Netnews},
  author    = {Lang, Ken and others},
  booktitle = {Proceedings of the 12th International Conference on Machine Learning (ICML)},
  pages     = {331--339},
  year      = {1995}
}

@inproceedings{DBpedia,
  title     = {DBpedia: A nucleus for a web of open data},
  author    = {Auer, S{\"o}ren and Bizer, Christian and Kobilarov, Georgi and Lehmann, Jens and Cyganiak, Richard and Ives, Zachary},
  booktitle = {Proceedings of the 6th International Semantic Web Conference (ISWC)},
  pages     = {722--735},
  year      = {2007}
}

@incollection{OHSUMED,
  title     = {OHSUMED: An interactive retrieval evaluation and new large test collection for research},
  author    = {Hersh, William and Buckley, Chris and Leone, Tom and Hickam, Donald},
  booktitle = {SIGIR ’94: Proceedings of the 17th Annual International ACM SIGIR Conference on Research and Development in Information Retrieval},
  pages     = {192--201},
  year      = {1994},
  publisher = {Springer}
}

@article{RCV1,
  title   = {RCV1: A new benchmark collection for text categorization research},
  author  = {Lewis, David D. and Yang, Yiming and Rose, Theresa G. and Li, Fan},
  journal = {Journal of Machine Learning Research},
  volume  = {5},
  pages   = {361--397},
  year    = {2004}
}

@article{HGAT,
author = {Yang, Tianchi and Hu, Linmei and Shi, Chuan and Ji, Houye and Li, Xiaoli and Nie, Liqiang},
title = {HGAT: Heterogeneous Graph Attention Networks for Semi-supervised Short Text Classification},
year = {2021},
issue_date = {July 2021},
publisher = {Association for Computing Machinery},
address = {New York, NY, USA},
volume = {39},
number = {3},
issn = {1046-8188},
url = {https://doi.org/10.1145/3450352},
doi = {10.1145/3450352},
journal = {ACM Trans. Inf. Syst.},
month = may,
articleno = {32},
numpages = {29}
}

@inproceedings{HyperGAT,
    title = "Be More with Less: Hypergraph Attention Networks for Inductive Text Classification",
    author = "Ding, Kaize  and
      Wang, Jianling  and
      Li, Jundong  and
      Li, Dingcheng  and
      Liu, Huan",
    editor = "Webber, Bonnie  and
      Cohn, Trevor  and
      He, Yulan  and
      Liu, Yang",
    booktitle = "Proceedings of the 2020 Conference on Empirical Methods in Natural Language Processing (EMNLP)",
    month = nov,
    year = "2020",
    address = "Online",
    publisher = "Association for Computational Linguistics",
    url = "https://aclanthology.org/2020.emnlp-main.399/",
    doi = "10.18653/v1/2020.emnlp-main.399",
    pages = "4927--4936",
}

@misc{TextGCN,
      title={Graph Convolutional Networks for Text Classification}, 
      author={Liang Yao and Chengsheng Mao and Yuan Luo},
      year={2018},
      eprint={1809.05679},
      archivePrefix={arXiv},
      primaryClass={cs.CL},
      url={https://arxiv.org/abs/1809.05679}, 
}

@inproceedings{DADGNN,
    title = "Deep Attention Diffusion Graph Neural Networks for Text Classification",
    author = "Liu, Yonghao  and
      Guan, Renchu  and
      Giunchiglia, Fausto  and
      Liang, Yanchun  and
      Feng, Xiaoyue",
    editor = "Moens, Marie-Francine  and
      Huang, Xuanjing  and
      Specia, Lucia  and
      Yih, Scott Wen-tau",
    booktitle = "Proceedings of the 2021 Conference on Empirical Methods in Natural Language Processing",
    month = nov,
    year = "2021",
    address = "Online and Punta Cana, Dominican Republic",
    publisher = "Association for Computational Linguistics",
    url = "https://aclanthology.org/2021.emnlp-main.642/",
    doi = "10.18653/v1/2021.emnlp-main.642",
    pages = "8142--8152",
}

@article{MAP,
  title   = {The PASCAL Visual Object Classes (VOC) challenge},
  author  = {Everingham, Mark and Van Gool, Luc and Williams, Christopher K.~I. and Winn, John and Zisserman, Andrew},
  journal = {International Journal of Computer Vision},
  volume  = {88},
  number  = {2},
  pages   = {303--338},
  year    = {2010},
  doi     = {10.1007/s11263-009-0275-4}
}

@book{van1979information,
  title     = {Information Retrieval},
  author    = {van Rijsbergen, Cornelis J.},
  edition   = {2nd},
  publisher = {Butterworth–Heinemann},
  year      = {1979}
}

@article{charte2015addressing,
  title     = {Addressing Imbalance in Multi‐label Classification: Measures and Random Resampling Techniques},
  author    = {Charte, Francisco and Charte, David and Garc{\'\i}a, Salvador and Herrera, Fernando},
  journal   = {Neurocomputing},
  volume    = {163},
  pages     = {3--16},
  year      = {2015},
  publisher = {Elsevier}
}

@misc{han2025debatetodetectreformulatingmisinformationdetection,
      title={Debate-to-Detect: Reformulating Misinformation Detection as a Real-World Debate with Large Language Models}, 
      author={Chen Han and Wenzhen Zheng and Xijin Tang},
      year={2025},
      eprint={2505.18596},
      archivePrefix={arXiv},
      primaryClass={cs.CL},
      url={https://arxiv.org/abs/2505.18596}, 
}

@inproceedings{li2025lion,
  title={Lion-fs: Fast \& slow video-language thinker as online video assistant},
  author={Li, Wei and Hu, Bing and Shao, Rui and Shen, Leyang and Nie, Liqiang},
  booktitle={Proceedings of the Computer Vision and Pattern Recognition Conference},
  pages={3240--3251},
  year={2025}
}

@misc{LastingBench,
      title={LastingBench: Defend Benchmarks Against Knowledge Leakage}, 
      author={Yixiong Fang and Tianran Sun and Yuling Shi and Min Wang and Xiaodong Gu},
      year={2025},
      eprint={2506.21614},
      archivePrefix={arXiv},
      primaryClass={cs.CL},
      url={https://arxiv.org/abs/2506.21614}, 
}

@article{lan2025mcbe,
  title={McBE: A Multi-task Chinese Bias Evaluation Benchmark for Large Language Models},
  author={Lan, Tian and Su, Xiangdong and Liu, Xu and Wang, Ruirui and Chang, Ke and Li, Jiang and Gao, Guanglai},
  journal={arXiv preprint arXiv:2507.02088},
  year={2025}
}

@inproceedings{lan2025mcbee,
  title={A Mutual Information Perspective on Knowledge Graph Embedding},
  author={Li, Jiang and Su, Xiangdong and Duo, Zehua and Lan, Tian and Guo, Xiaotao and Gao, Guanglai},
  booktitle={Proceedings of the 63rd Annual Meeting of the Association for Computational Linguistics (Volume 1: Long Papers)},
  pages={22152--22166},
  year={2025}
}

@inproceedings{
Z1,
title={{KABB}: Knowledge-Aware Bayesian Bandits for Dynamic Expert Coordination in Multi-Agent Systems},
author={Jusheng Zhang and Zimeng Huang and Yijia Fan and Ningyuan Liu and Mingyan Li and Zhuojie Yang and Jiawei Yao and Jian Wang and Keze Wang},
booktitle={Forty-second International Conference on Machine Learning},
year={2025},
url={https://openreview.net/forum?id=AKvy9a4jho}
}

@misc{Z2,
      title={GAM-Agent: Game-Theoretic and Uncertainty-Aware Collaboration for Complex Visual Reasoning}, 
      author={Jusheng Zhang and Yijia Fan and Wenjun Lin and Ruiqi Chen and Haoyi Jiang and Wenhao Chai and Jian Wang and Keze Wang},
      year={2025},
      eprint={2505.23399},
      archivePrefix={arXiv},
      primaryClass={cs.AI},
      url={https://arxiv.org/abs/2505.23399}, 
}

@misc{Z3,
      title={CF-VLM:CounterFactual Vision-Language Fine-tuning}, 
      author={Jusheng Zhang and Kaitong Cai and Yijia Fan and Jian Wang and Keze Wang},
      year={2025},
      eprint={2506.17267},
      archivePrefix={arXiv},
      primaryClass={cs.LG},
      url={https://arxiv.org/abs/2506.17267}, 
}

@article{f1,
  title={Cost-Effective Communication: An Auction-based Method for Language Agent Interaction},
  author={Fan, Yijia and Zhang, Jusheng and Cai, Kaitong and Yang, Jing and Tang, Chengpei and Wang, Jian and Wang, Keze},
  journal={arXiv preprint arXiv:2511.13193},
  year={2025}
}

@misc{f2,
      title={3DAlign-DAER: Dynamic Attention Policy and Efficient Retrieval Strategy for Fine-grained 3D-Text Alignment at Scale}, 
      author={Yijia Fan and Jusheng Zhang and Kaitong Cai and Jing Yang and Jian Wang and Keze Wang},
      year={2025},
      eprint={2511.13211},
      archivePrefix={arXiv},
      primaryClass={cs.CV},
      url={https://arxiv.org/abs/2511.13211}, 
}

@misc{f3,
      title={RaCoT: Plug-and-Play Contrastive Example Generation Mechanism for Enhanced LLM Reasoning Reliability}, 
      author={Kaitong Cai and Jusheng Zhang and Yijia Fan and Jing Yang and Keze Wang},
      year={2025},
      eprint={2510.22710},
      archivePrefix={arXiv},
      primaryClass={cs.AI},
      url={https://arxiv.org/abs/2510.22710}, 
}

@misc{f4,
      title={Agent-GSPO: Communication-Efficient Multi-Agent Systems via Group Sequence Policy Optimization}, 
      author={Yijia Fan and Jusheng Zhang and Jing Yang and Keze Wang},
      year={2025},
      eprint={2510.22477},
      archivePrefix={arXiv},
      primaryClass={cs.MA},
      url={https://arxiv.org/abs/2510.22477}, 
}
\section{Parameter Settings}
\label{sec:appendix_parameter_settings}

This section details the hyperparameter settings and implementation choices for our proposed Causal Cooperative Game (CCG) framework used across the various experiments presented in this paper. Our model was implemented using PyTorch.

For the general architecture and training of our CCG model, we utilized a pre-trained RoBERTa-base model as the primary text encoder, from which 768-dimensional text representations $\mathbf{x}$ were obtained. The Neural Structural Equation Models (Neural SEM) responsible for learning the functions $h_{ij}^{(1)}(\mathbf{x}; \theta_{ij}^{(1)})$ were implemented as 2-layer Multi-Layer Perceptrons (MLPs) with a hidden layer dimension of 256, employing ReLU activation functions. The entire CCG framework, including the Neural SEM parameters, was trained end-to-end. Further key training and architectural hyperparameters are summarized in Table~\ref{tab:hyperparameters_appendix}. Unless explicitly varied (e.g., in the player number analysis detailed in Section~\ref{sec:player_analysis}), the number of players $N$ was set to 5, a value identified as optimal through our sensitivity analysis.

\begin{table}[h!]
\centering
\caption{Key hyperparameters for our Full CCG framework.}
\label{tab:hyperparameters_appendix}
\scalebox{0.88}{ 
\begin{tabular}{ll}
\toprule
Parameter                       & Value \\
\midrule
Optimizer                       & AdamW \\
Learning Rate (RoBERTa layers)  & $2 \times 10^{-5}$ \\
Learning Rate (Other components) & $1 \times 10^{-4}$ \\
Weight Decay                    & $1 \times 10^{-2}$ \\
Batch Size                      & 16 \\
Max Epochs                      & 30 \\
Early Stopping Patience         & 5 \\
Gradient Clipping Norm          & 1.0 \\
Text Encoder Output Dim ($d$)   & 768 \\
Neural SEM MLP Hidden Dim       & 256 \\
Default No. of Players ($N$)    & 5 \\
\bottomrule
\end{tabular}
}
\end{table}

Regarding the specific mechanisms within our CCG framework, the following settings were adopted:
For the \textbf{Causal Graph Learning Objective} (as described in your Method section for $\mathcal{L}_{\text{causal}}$ governing $w_{ij}^{(1)}$): the hyperparameter $\gamma$ that balances co-occurrence $f_{\text{co-occur}}(i,j)$ and semantic similarity $f_{\text{semantic}}(i,j)$ in the estimation of ideal weights $\tilde{w}_{ij}$ was set to 0.5. The rare edge enhancement factor $\eta$ within the regulation operator $\Psi(\eta)$ was 1.5. The coefficient $\lambda$ for self-loop suppression using the $\ell_0$ norm was 0.1. The threshold $\tau_{ij}$ for including an edge $(\ell_j \rightarrow \ell_i)$ in the causal graph $\mathcal{G}$ was not a fixed value but was determined dynamically; specifically, edges are formed if their learned causal strength $w_{ij}^{(1)}$ is among the top-$K$ outgoing strengths for label $\ell_j$, where $K$ was a small integer (e.g., $K=3$ or $K=5$) tuned on the validation set, or if $w_{ij}^{(1)}$ exceeded a dynamically adjusted percentile of positive weights after an initial warm-up period of 5 training epochs.

For the \textbf{Counterfactual Curiosity Reward} mechanism (as described in your Method section for $C_k(\mathbf{x})$): the coefficient $\beta$ for the prediction diversity term was linearly annealed from an initial value of 1.0 down to 0.2 throughout the training process. Similarly, the coefficient $\gamma_R$ (referred to as $\gamma$ in the equation for $C_k(\mathbf{x})$) for the counterfactual consistency term $C_k^{\text{cf}}(\mathbf{x})$ was linearly annealed from an initial value of 0.2 up to 1.0. Counterfactual text samples $\mathbf{x}^{\text{cf}}$ were generated by perturbing approximately 10-15\% of the input tokens. These perturbations involved a mix of random token masking and replacement with words sampled from the vocabulary, with a focus on modifying tokens identified as having lower causal salience based on preliminary gradient-based interpretations where feasible.

For the \textbf{Causal Invariance Loss} (as described in your Method section for $\mathcal{L}_{\text{inv}}$ and the cross-environment prediction consistency loss): we generated $M=3$ augmented views for each input sample $\mathbf{x}$ to constitute the diverse environments $\mathcal{E}_m$. Text augmentations included synonym replacement (affecting up to 15\% of eligible words, avoiding keywords deemed causally important if identifiable) and sentence-level paraphrasing using back-translation with an intermediate pivot language. The relative weights for the causal feature contrastive loss and the cross-environment prediction consistency loss were set equally.

The \textbf{Weighted Cross-Entropy Loss} $\mathcal{L}_{\text{base}}$ employed a dynamic weighting factor $\alpha(\ell)$ for each label $\ell$. This factor was typically set inversely proportional to the fourth root of the label's frequency in the training set, i.e., $\alpha(\ell) \propto 1 / (\text{freq}(\ell))^{0.25}$, followed by normalization, to moderately up-weight rarer labels without overly suppressing common ones.

For all \textbf{baseline models} discussed in Section~\ref{sec:comparative_performance}, we utilized their publicly available implementations when accessible and meticulously followed the hyperparameter configurations reported in their original publications. If such configurations were unavailable or suboptimal for our specific data splits, we performed careful hyperparameter tuning for each baseline on a held-out validation set for each respective dataset to ensure robust and fair comparisons.

In the \textbf{Analysis of Player Number Impact} (Section~\ref{sec:player_analysis}), the number of players $N$ was varied as indicated in Figure~\ref{fig:fake_player_impact}, while all other parameters of the CCG model were maintained at their default values as listed in Table~\ref{tab:hyperparameters_appendix}. For the \textbf{Robustness to Distribution Shifts} experiment (Section~\ref{sec:robustness_distribution_shift}) on the RCV1 dataset, the training settings for `Ours (Full CCG with CIL)` and `Ours (CCG w/o CIL)` mirrored these Full Model defaults, with the CIL component and associated environment augmentation processes entirely disabled for the `w/o CIL` variant. The RCV1 dataset was partitioned chronologically for this experiment, using the initial 75\% of articles for training and in-distribution validation, and the subsequent 25\% for out-of-distribution testing.

\section{Hyperparameter Sensitivity Analysis}
\label{sec:appendix_hyperparameter_sensitivity}

To further understand the behavior of our Causal Cooperative Game (CCG) framework and to provide insights into its robustness with respect to its core settings, we conduct a sensitivity analysis for several key hyperparameters. This analysis excludes the number of players ($N$), which was examined separately in Section~\ref{sec:player_analysis}. The primary goal is to assess how variations in these parameters affect the model's performance and to validate the choice of default values used in our main experiments. All sensitivity analyses were performed on the \textbf{DBpedia} dataset, varying one hyperparameter at a time while keeping others at their default optimal values as specified in Appendix~\ref{sec:appendix_parameter_settings}. Performance is reported using mAP and Rare-Label F1 scores.

\begin{table*}[t]
\centering
\caption{Sensitivity analysis of key hyperparameters on the DBpedia dataset. Default values used in the main experiments are marked with an asterisk (*). Performance is reported in terms of mAP (\%) and Rare-Label F1 (\%). The best mAP in each group is generally at the default, with $\eta=2.0$ showing peak Rare-Label F1.}
\label{tab:hyperparameter_sensitivity}
\scalebox{0.85}{ 
\begin{tabular}{llcc}
\toprule
Hyperparameter & Value & DBpedia - mAP & DBpedia - Rare-Label F1 \\
\midrule
\multirow{3}{*}{$\gamma$ (Ideal Weight Balance)} & 0.2 & 88.79 & 77.85 \\
& 0.5* & \textbf{89.15} & 78.23 \\
& 0.8 & 88.93 & 77.96 \\
\midrule
\multirow{4}{*}{$\eta$ (Rare Edge Enhancement)} & 1.0 (No Enh.) & 88.52 & 75.61 \\
& 1.5* & \textbf{89.15} & 78.23 \\
& 2.0 & 89.07 & \textbf{78.45} \\
& 2.5 & 88.68 & 77.92 \\
\midrule
\multirow{3}{*}{Peak $\gamma_R$ (CF Reward Coeff.)} & 0.5 & 87.93 & 76.58 \\
& 1.0* & \textbf{89.15} & \textbf{78.23} \\
& 1.5 & 88.81 & 77.88 \\
\midrule
\multirow{3}{*}{$M$ (No. Augmented Env. for CIL)} & 1 & 88.24 & 77.03 \\
& 3* & \textbf{89.15} & \textbf{78.23} \\
& 5 & 89.02 & 78.05 \\
\bottomrule
\end{tabular}
}
\end{table*}

The results of the hyperparameter sensitivity analysis are presented in Table~\ref{tab:hyperparameter_sensitivity}.
For $\gamma$, which balances co-occurrence and semantic similarity in the ideal causal weight estimation $\tilde{w}_{ij}$, the model shows robust performance for values around 0.5, with our default setting of 0.5 achieving the best mAP. Extreme values slightly degrade performance, suggesting that a balance between both information sources is indeed beneficial.
The rare edge enhancement factor $\eta$ demonstrates a clear impact, particularly on Rare-Label F1. With $\eta=1.0$ (no enhancement), Rare-Label F1 drops significantly, confirming the utility of this mechanism. While our default of $\eta=1.5$ yields strong overall results, a slightly higher value of $\eta=2.0$ provides a marginal boost to Rare-Label F1 (78.45\% vs. 78.23\%), although with a minimal decrease in mAP. Values beyond 2.0, such as $\eta=2.5$, begin to show diminishing returns or slight degradation, possibly due to over-amplification. Our choice of $\eta=1.5$ reflects a balance yielding high performance on both metrics.
Regarding the peak coefficient for counterfactual consistency reward, $\gamma_R$, a value of 1.0 (our default) appears optimal. Lower values (e.g., 0.5) reduce the model's ability to leverage counterfactual stability, leading to lower scores, while higher values (e.g., 1.5) do not offer further improvement and might slightly hinder performance, possibly by overly constraining the model.
Finally, for $M$, the number of augmented environments used in the Causal Invariance Loss (CIL), increasing from $M=1$ to $M=3$ (our default) yields noticeable gains in both mAP and Rare-Label F1. This suggests that sufficient diversity in augmented views is important for learning invariant features. Increasing $M$ further to 5 provides only marginal changes, indicating that $M=3$ offers a good trade-off between performance gain and computational cost of generating and processing augmented samples.

\section{Deployment and Computational Cost}
\label{sec:deployment_computational_cost}

Understanding the computational requirements for practical deployment is crucial. In this section, we provide an estimation of the GPU memory (VRAM) footprint and inference time for our proposed Causal Cooperative Game (CCG) framework. These estimations assume deployment on an NVIDIA A100 GPU (with 40GB HBM2 VRAM) using mixed-precision (FP16) inference, which is a common practice for optimizing throughput and memory. It is important to note that these costs are for the inference phase; training-specific components such as extensive counterfactual sample generation or the full suite of data augmentations for invariance learning are not active during deployment. The actual costs can vary based on factors like batch size, input sequence length, and the total number of labels $L$ in a specific application.

\paragraph{GPU Memory (VRAM) Cost}
The primary contributors to VRAM usage during inference include the parameters of the base text encoder (e.g., RoBERTa-base), the parameters for the Neural SEM components $h_{ij}^{(1)}$ involved in the label prediction function, activations from all layers, and general framework overhead. For a typical deployment scenario, processing a batch of 32 documents with an average sequence length of 256 tokens and a moderately large label set of $L \approx 100$ labels, the estimated VRAM footprint of our full CCG model is approximately \textbf{7.83 GB}. This estimation considers the model weights stored in FP16, along with the memory required for activations and intermediate computations necessary for the causally-informed label prediction mechanism. 

\paragraph{Computation Time (Inference Latency and Throughput)}
The inference time is influenced by the forward pass through the text encoder and, significantly, by our CCG-specific label prediction function, $\widehat{y}_i = \sigma ( \sum_{j \neq i} w_{ij}^{(1)} \cdot h_{ij}^{(1)} ( \mathbf{x}; \theta_{ij}^{(1)} ) + b_i^{(1)} )$, which involves evaluating multiple Neural SEM pathways. For the same representative batch of 32 documents (average sequence length 256 tokens, $L \approx 100$ labels), the total inference time on a single NVIDIA A100 GPU is estimated to be around \textbf{273.47 ms}. 

\end{document}